\documentclass[conference,10pt]{IEEEtran}

\usepackage{graphicx,bm}
\usepackage{epstopdf}
\usepackage{amssymb}
\usepackage{todonotes}
\usepackage[utf8]{inputenc}
\usepackage{amsmath}
\usepackage{esvect}
\let\labelindent\relax
\usepackage{enumitem}
\renewcommand{\(}{\left(}
\renewcommand{\)}{\right)}
\renewcommand{\[}{\left[}
\renewcommand{\]}{\right]}

\renewcommand{\b}{\mathbf{b}}

\renewcommand{\L}{\mathbf{L}}

\newcommand{\thet}{\bm{\theta}}

\newcommand{\y}{\mathbf{y}}
\renewcommand{\H }{\mathbf{H}}
\newcommand{\z}{\mathbf{z}}
\newcommand{\w}{\mathbf{w}}

\newcommand{\W}{\mathbf{W}}

\newcommand{\0}{\mathbf{0}}

\newcommand{\x}{\mathbf{x}}

\renewcommand{\t}{\mathbf{t}}
\newcommand{\A}{\mathbf{A}}

\renewcommand{\v}{\mathbf{v}}

\renewcommand{\b}{\mathbf{b}}

\newcommand{\EE}[1]{{\rm{E}}\left\{#1\right\}}

\renewcommand{\log}[1]{{\rm{log}}#1}
\renewcommand{\arg}[1]{{\rm{arg}}#1}

\begin{document}
\title{Deep MIMO Detection}

\author{Neev Samuel, Tzvi Diskin and Ami Wiesel\\\\School of Computer Science and Engineering\\
The Hebrew University of Jerusalem\\
The Edmond J. Safra Campus\\
9190416 Jerusalem, Israel}

\maketitle

\begin{abstract}
In this paper, we consider the use of deep neural networks in the context of Multiple-Input-Multiple-Output (MIMO) detection. We give a brief introduction to deep learning and propose a modern neural network architecture suitable for this detection task. First, we consider the case in which the MIMO channel is constant, and we learn a detector for a specific system. Next, we consider the harder case in which the parameters are known yet changing and a single detector must be learned for all multiple varying channels. We demonstrate the performance of our deep MIMO detector using numerical simulations in comparison to competing methods including approximate message passing and semidefinite relaxation. The results show that deep networks can achieve state of the art accuracy with significantly lower complexity while providing robustness against ill conditioned channels and mis-specified noise variance. 
\end{abstract}

\begin{IEEEkeywords}
MIMO Detection, Deep Learning, Neural Networks.
\end{IEEEkeywords}

\section{Introduction}
Multiple input multiple output (MIMO) systems arise in most modern communication channels. The dimensions can account for time and frequency resources, multiple users, multiple antennas and other resources. These promise substantial
performance gains, but present a challenging detection problem in terms of computational complexity.
In recent years, the world is witnessing a revolution in deep machine learning. In many fields of engineering, e.g., computer vision, it was shown that computers can be fed with sample pairs of inputs and desired outputs, and ``learn'' the functions which relates them. These rules can then be used to classify (detect) the unknown outputs of future inputs. The goal of this paper is to apply deep machine learning in the classical MIMO detection problem and understand its advantages and disadvantages.

\subsection{Background on MIMO detection}
The binary MIMO detection setting is a classical problem in simple hypothesis testing \cite{verdu1998multiuser}. The maximum
likelihood (ML) detector is 
the optimal detector in the sense of minimum joint probability of
error for detecting all the symbols simultaneously.
It can be implemented via efficient search algorithms, e.g., the
sphere decoder \cite{agrell2002closest}. The difficulty is that its worst case computational complexity is impractical for many applications.
Consequently, several modified search algorithms have been purposed, offering improved complexity performance \cite{guo2006algorithm}\cite{suh2017reduced}.
There has been much interest in implementing
suboptimal detection algorithms. The most common suboptimal
detectors are the linear receivers, i.e., the matched filter (MF),
the decorrelator or zero forcing (ZF) detector and the minimum
mean squared error (MMSE) detector. More advanced detectors are
based on decision feedback equalization (DFE), approximate message passing (AMP) \cite{jeon2015optimality} and semidefinite relaxation (SDR) \cite{luo2010semidefinite,jalden2008diversity}. Currently, both AMP and SDR provide near optimal accuracy under many practical scenarios. AMP is simple and cheap to implement in practice, but is an iterative method that may diverge in problematic settings. SDR is more robust and has polynomial complexity, but is much slower in practice.

\subsection{Background on Machine Learning}
In the last decade, there is an explosion of machine learning success stories in all fields of engineering. Supervised classification is similar to statistical detection theory. Both observe noisy data and output a decision on the discrete unknown it originated from. Typically, the two fields differ in that detection theory is based on a prior probabilistic model of the environment, whereas learning is data driven and is based on examples. In the context of MIMO detection, a model is known and allows us to generate as many synthetic examples as needed. Therefore we adapt an alternative notion. We interpret ``learning'' as the idea of choosing a best decoder from a prescribed class of algorithms. Classical detection theory tries to choose the best estimate of the unknowns, whereas machine learning tries to choose the best algorithm to be applied. Indeed, the hypotheses in detection are the unknown symbols, whereas the hypotheses in learning are the detection rules \cite{Shalev}. Practically, this means that the computationally involved part of detection is applied every time we get a new observation. In learning, the expensive stage is learning the algorithm which is typically performed off line. Once the optimal rule algorithm is found, we can cheaply implement it in real time. 

Machine learning has a long history but was previously limited to simple and small problems. Fast forwarding to the last years, the field witnessed the deep revolution. The ``deep'' adjective is associated with the use of complicated and expressive classes of algorithms, also known as architectures. These are typically neural networks with many non-linear operations and layers. Deep architectures are more expressive than shallow ones \cite{lecun2015deep}, but were previously considered impossible to optimize. With the advances in big data, optimization algorithms and stronger computing resources, such networks are currently state of the art in different problems including speech processing and computer vision. 
In particular, one promising approach to designing deep architectures is by unfolding an existing iterative algorithm \cite{hershey2014deep}. Each iteration is considered a layer and the algorithm is called a network. The learning begins with the existing algorithm as an initial starting point and uses optimization methods to improve the algorithm. For example, this strategy has been shown successful in the context of sparse reconstruction. Leading algorithms as Iterative Shrinkage and Thresholding and a sparse version of AMP have both been improved by unfolding their iterations into a network and learning their optimal parameters \cite{gregor2010learning,borgerding2016onsager}. 

In recent years, deep learning methods have been purposed for improving the performance of a decoder for linear codes in fixed channels\cite{nachmani2016learning}.
And in \cite{o2017introduction} several applications of deep learning for communication applications have been considered, including decoding signals over fading channels, but the architecture purposed there does not seem to be scalable for higher dimension signals.

\subsection{Main contributions}
The main contribution of this paper is the introduction of DetNET, a deep learning network for MIMO detection. DetNet is derived by unfolding a projected gradient descent method. Simulations show that it achieves near optimal detection performance while being a fast algorithm that can be implemented in real-time. Its accuracy is similar to SDR with running time that is more than 30 times faster. Compared to AMP, another detector with optimality guarantees, DetNet is more robust. It shows promising performance in handling ill conditioned channels, and does not require knowledge of the noise variance. 

Another important contribution, in the general context of deep learning, is DetNet's ability to perform on multiple models with a single training. Recently, there were many works on learning to invert linear channels and reconstruct signals \cite{gregor2010learning,borgerding2016onsager,Mousavi2017Learning}. To the best of our knowledge, all of these were developed and trained to address a single fixed channel. In contrast, DetNet is designed for handling multiple channels simultaneously with a single training phase.

\subsection{Notation}
In this paper, we shall define the normal distribution where $\mu$ is the mean and $\sigma^{2}$ is the variance as $\mathcal{N}\(\mu,\sigma^{2}\)$.  The uniform distribution with the minimum value $a$ and the maximum value $b$ will be $\mathcal{U}\(a,b\)$ .
Boldface uppercase letters denote matrices, Boldface lowercase letters denote vectors, the superscript $\(\cdot\)^{T}$ denotes the transpose. The i'th element of the vector $\x$ will be denoted as $\x_{i}$.
Unless stated otherwise, the term independent and identically distributed  (i.i.d.) Gaussian matrix, will refer to a matrix where each of its elements is i.i.d. sampled from the normal distribution $\mathcal{N}\(0,1\)$.
The rectified linear unit defined as $\rho(x) = \max\{0,x\}$ will be denoted as $\rho$.

\section{Learning to detect}
In this section, we formulate the MIMO detection problem in a machine learning framework.
We consider the standard linear MIMO model:
\begin{equation}\label{linearmodel2}
 \y = \H\x + \w,
\end{equation}
where $\y\in \mathbb{R}^{N}$ is the received vector, $\H\in \mathbb{R}^{N \times K}$ is the channel matrix, $\x\in \{\pm 1\}^{K}$ is an unknown vector of independent and equal probability binary symbols,
$\w\in \mathbb{R}^{N}$ is a noise vector with independent, zero mean Gaussian variables of variance $\sigma^2$. We do not assume knowledge of the variance as hypothesis testing theory guarantees that this is unnecessary for optimal detection. Indeed, the optimal ML rule does not require knowledge of $\sigma^2$. This is contrast to the MMSE and AMP decoders that exploit this parameter and are therefore less robust.

We assume perfect channel state information (CSI) and that the channel $\H$ is exactly known. However, we differentiate between two possible cases:
\begin{itemize}
 \item Fixed Channel (FC): In the FC scenario, $\H$ is deterministic and constant (or a realization of a degenerate distribution which only takes a single value). 
\item Varying Channel (VC): In the VC scenario, we assume $\H$ random with a known distribution.
\end{itemize}

Our goal is to detect $\x$, using an algorithm that receives $\y$ and $\H$ as inputs and estimates $\hat{\x}$.

The first step is choosing and fixing a detection architecture. 
An architecture is a function $\hat{\x}_{\thet}(\H,\y)$ that detects the unknown symbols given $\y$ and $\H$. The architecture is parametrized by $\thet$.
Learning is the problem of finding the $\thet$ within the feasible set that will lead to strong detectors $\hat{\x}_{\thet}(\H,\y)$.
By choosing different functions and parameter sets, we characterize competing types of detectors which tradeoff accuracy with complexity. 

To find the best detector, we fix a loss function $l\(\x;\hat{\x}_{\thet}\(\H,\y\)\)$ that measures the distance between the true symbols and their estimates. 
Then, we find $\thet$ by minimizing the loss function we chose over the MIMO model distribution:
\begin{eqnarray}\label{learning_min}
 \min_{\thet} \EE{ l\(\x;\hat{\x}_{\thet}(\H,\y)\)},
\end{eqnarray}
where the expectation is with respect to all the random variables in (\ref{linearmodel2}), i.e., $\x$,  $\w$, and $\H$. Learning to detect is defined as finding the best set of parameters $\thet$ of the architecture $\hat{\x}_{\thet}\(\y,\H\)$ that minimize the expected loss $l\(\cdot;\cdot\)$ over the distribution in (\ref{linearmodel2}).

The next examples illustrate how the choice of architecture $\hat{\x}_{\thet}\(\y,\H\)$ leads to different detectors that tradeoff accuracy for complexity.

{\em{Example 1:}} The goal in detection is to decrease the probability of error. 
Therefore, the best loss function in this problem 
\begin{eqnarray}
 l\(\x;\hat{\x}_{\thet}(\H,\y)\)=\left\{\begin{array}{ll}
    1  & \x\neq {\hat{\x}_{\thet}\(\y,\H\)} \\
    0  & {\rm{else}}.
 \end{array}\right.
\end{eqnarray}
By choosing an unrealistically flexible architecture with unbounded parameterization and no restrictions such that 
\begin{eqnarray}\label{abstractThet}
 \left\{\hat{\x}_{\thet}\(\y,\H\)\;:\;\thet\right\}=\left\{
\begin{array}{l}
    {\rm{all\;the\; functions\;}}  \\
   \mathbb{R}^{N}\times \mathbb{R}^{N \times K} \mapsto \{\pm 1\}^{K}
\end{array} 
\right \}.
\end{eqnarray}
Then, the solution to (\ref{learning_min}) is the ML decoder:
\begin{eqnarray}\label{mle}
 \hat{\x}_{\thet}\(\y,\H\)=\arg\min_{\x\in\{\pm 1\}^K}\|\y-\H\x\|^2.
\end{eqnarray}
This rule is optimal in terms of accuracy but requires a computationally intensive search of $O\(2^K\)$. 
Obviously, this example is theoretical since the architecture of all possible functions cannot be  parametrized and (\ref{learning_min}) cannot be optimized.

{\em{Example 2}}: On the other extreme, consider the architecture of fixed linear detectors:
\begin{eqnarray}\label{hc2}
 \hat{\x}_{\thet}\(\y,\H\)=\A\y,
\end{eqnarray}
where the parameter $\thet$ is a single fixed matrix to be optimized within ${\mathbb{R}}^{K\times N}$.
In the FC model, choosing $\|\x- \hat{\x}(\y,\H)\|^2$ as the loss function and assuming $\sigma^2\rightarrow 0$, the optimal decoder is the well known decorrelator:
\begin{eqnarray}\label{ls}
 \hat{\x}_{\thet}\(\y,\H\)=\(\H^T\H\)^{-1}\H^T\y.
\end{eqnarray}
The resulting detector involves a simple matrix multiplication that requires $O(NK)$ operations, but is not very accurate.
On the other hand, if we consider the more challenging VC model, then the optimal linear transformation is simply $\H=\0$. A single linear decoder cannot decode arbitrary channels simultaneously, and the decoder is completely useless. 

These two examples emphasize how fixing an architecture and a loss function determines what will be the optimal detector for the MIMO detection problem. The more expressive we choose $\hat{\x}$ to be, the more accurate the final detector can be, on the expense of the computational complexity. 


We close this section with a technical note on the numerical implementation of the optimization in (\ref{learning_min}).
In practice, it is intractable to optimize over an abstract class of functions as in (\ref{abstractThet}). Numerical minimization is typically performed with respect to a finite parameter set as in (\ref{hc2}).  Thus, our deep architectures are based on multiple layers with multivariate linear operations and element-wise non-linear operators. These allow rich decoding functions while resorting to a finite and tractable parameterization. In addition,
analytic computation of the expectation in the objective is usually impossible. Instead, we approximate it using an empirical mean of samples drawn from a data set of examples (thus the 'learning' notion). In our case, the data set is composed of synthetically generated samples satisfying (\ref{linearmodel2}). Both these technicalities, were considered unthinkable just a decade ago, but are now standard procedures in the deep learning community. Easy to use, open source tools, make it possible to create deep architectures and optimize them in a straight forward manner. Specifically, in this work, all the experiments were implemented on the TensorFlow framework \cite{abadi2016tensorflow}.

\section{Deep MIMO detector}

\begin{figure}[t]
  \centering
  \center{\includegraphics[width=8.7cm]{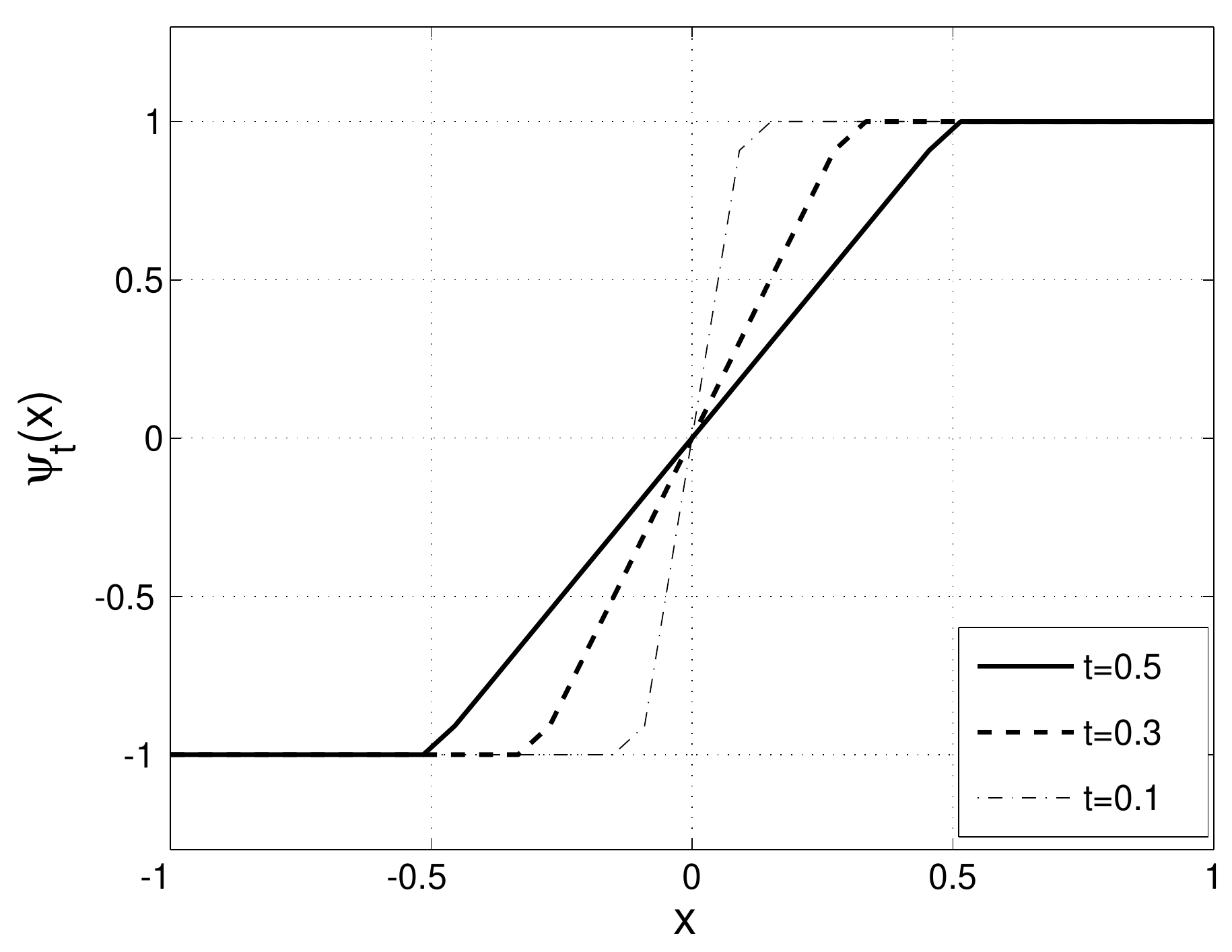}}
%
\caption{A graph illustrating the linear soft sign function $\psi_{t}(\x)$ for different values of the parameter $t$.}
\label{fig:soft_sign}
\end{figure}

\begin{figure}[t]
  \center{\includegraphics[width=8.9cm]{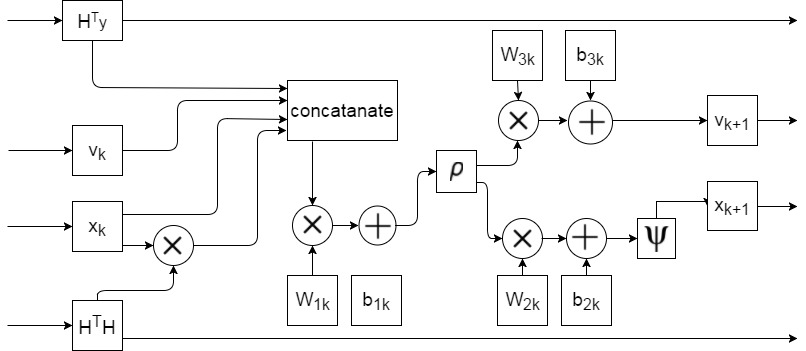}}
\caption{A flowchart representing a single layer of DetNet.}
\label{fig:DetNet_arch}%
\end{figure}

In this section, we propose a deep detector with an architecture which is specifically designed for MIMO detection that will be named from now on 'DetNet' (Detection Network). First, we note that an efficient detector should not work with $\y$ directly, but use the compressed sufficient statistic:
\begin{eqnarray}
 \H^T\y=\H^T\H\x+\H^T\w.
\end{eqnarray}
This hints that two main ingredients in the architecture should be $\H^T\y$ and $\H^T\H\x$.  Second, our construction is based on mimicking a projected gradient descent like solution for the ML optimization in (\ref{mle}). Such an algorithm would lead to iterations of the form

\begin{eqnarray}
 \hat\x_{k+1}&=&\Pi\[\hat\x_k-\delta_k\left.\frac{\partial \|\y-\H\x\|^2}{\partial \x}\right|_{\x=\hat\x_k}\]\nonumber\\
 &=&\Pi\[\hat\x_k-\delta_k\H^T\y+\delta_k\H^T\H\x_k\],
\end{eqnarray}
 where $\hat\x_k$ is the estimate in the $k$'th iteration, $\Pi[\cdot]$ is a nonlinear projection operator, and $\delta_k$ is a step size. Intuitively, each iteration is a linear combination of the the $\x_k$, $\H^T\y$, and $\H^T\H\x_k$ followed by a non-linear projection. We enrich these iterations by lifting the input to a higher dimension and applying standard non-linearities which are common in deep neural networks. This yields the following architecture:
\begin{eqnarray}\label{DetNetArchitecture}
\nonumber\\
 \z_{k} &=& \rho\(\W_{1k} \[
 \begin{array}{c}
   \H^{T}\y \\   \hat\x_{k} \\ \H^{T} \H \hat \x_{k} \\ \v_{k} 
 \end{array} \] 
 +\b_{1k}\) \nonumber\\
 \hat\x_{k+1}&=&\psi_{t_k}\(\W_{2k}\z_k+\b_{2k}\)\nonumber\\
 \hat\v_{k+1}&=&\W_{3k}\z_k+\b_{3k} \nonumber\\
 \hat\x_1 &=& \0,
\end{eqnarray}
where $k=1,\cdots,L$ and $\psi_t(\cdot)$ is a piecewise linear soft sign operator defined as:
\begin{eqnarray}
\psi_t(x) =  -1+\frac{\rho(x+t)}{|t|}-\frac{\rho(x-t)}{|t|}.
\end{eqnarray}
The operator is plotted in Fig. \ref{fig:soft_sign}, and the structure of each DetNet layer is illustrated in Fig. \ref{fig:DetNet_arch}. 
The final estimate is defined as $\hat\x_{\thet}\(\y,\H\)=sign(\hat\x_L)$.

 
The parameters of DetNet that are optimized during the learning phase are:
\begin{eqnarray}
 \thet=\left\{\W_{1k},\b_{1k},\W_{2k},\b_{2k},\W_{3k},\b_{1k},\t_{k}\right\}_{k=1}^L.
\end{eqnarray}

Training  deep networks is a difficult task due to vanishing gradients, saturation of the activation functions, sensitivity to initializations and more \cite{glorot2010understanding}. To address these challenges, we adopted a loss function that takes into account the outputs of all of the layers.
Moreover, since the errors depend on the channel's realization, we decided to normalize the errors with those of the decorrelator. Together, this led to the following loss function:
\begin{eqnarray}
  l\(\x;\hat{\x}_{\thet}\(\H,\y\)\) = \sum_{k=1}^{\L} \log(k)
\frac{\|\x-\hat{\x}_{k}\|^2}{\|\x-\tilde\x\|^2},
\end{eqnarray}
where:
\begin{eqnarray}\label{decorrelator}
\tilde\x=\(\H^T\H\)^{-1}\H^T\y.
\end{eqnarray}
is the standard decorrelator decoder.

In our final implementation, in order to further enhance the performance of DetNet, we added a residual feature from ResNet \cite{he2016deep} where the output of each layer is a weighted average with the output of the previous layer. Note also that our loss function is motivated by the auxiliary classifiers feature in GoogLeNet \cite{szegedy2015going}.

We train the network using a variant of the stochastic gradient descent method \cite{rumelhart1988learning},\cite{bottou2010large} for optimizing deep networks, named Adam Optimizer \cite{kingma2014adam}.
We used batch training with $5000$  random data samples at each iteration, and trained the network for $50000$ iterations. 
To give a rough idea of the complexity, learning the detectors in our numerical results took 2 days on a standard Intel i7-6700 processor. Each sample was independently generated from (\ref{linearmodel2}) according to the statistics of $\x$, $\H$ (either in the FC or VC model) and $\w$. With respect to the noise, its variance is unknown and therefore this too was randomly generated so that the SNR will be uniformly distributed on $\mathcal{U}\({\rm{SNR}}_{\min},{\rm{SNR}}_{\max}\)$. 
This approach allows us to train the network to detect over a wide range of SNR values.

\section{Numerical results}
In this section, we demonstrate the advantages of our proposed detector using computer simulations. 

All the experiments address a MIMO channel with an input of size $K=30$ and output of size $N=60$. It is well known that performance is highly dependent on the type of MIMO channel. Therefore, we tried two scenarios:
\begin{description}
 \item[FC:] In this model, we chose to test the  algorithms on a deterministic and constant ill-conditioned matrix which is known to be challenging for detection \cite{rangan2014convergence}. The matrix was generated such that $\H^T\H$ would have a Toeplitz structure with $\[\H^T\H\]_{i,j}=0.55^{|i-j|}$. We shall denote this matrix as the 0.55-Toeplitz matrix. This defines the singular values and right singular vectors of $\H$. Its left singular vectors were randomly generated uniformly in the space of orthogonal matrices, and then fixed throughout the simulations. 
 \item[VC:] In this model, the matrices $\H$ were randomly generated with i.i.d. ${\mathcal{N}}\(0,1\)$ elements. Each example was independently generated within the same experiment.
\end{description}
We have tested the performance of the following detection algorithms:
\begin{description}
\item  [FCDN:] DetNet algorithm described in (\ref{DetNetArchitecture}) with $3K$ layers, $\z_k$ of size $8K$, and $\v_{k}$ of size $2K$. FCDN was trained using the FC model described above, and is specifically designed to handle a specific ill conditioned channel matrix.
 \item  [VCDN:] Same architecture as the FCDN but the training is on the VC model and is supposed to cope with arbitrary channel matrices.
 \item  [ShVCDN :] Same as the VCDN algorithm, but with a shallow network architecture using only $K$ layers.
 \item [ZF:] This is the classical decorrelator, also known as least squares or zero forcing (ZF) detector \cite{verdu1998multiuser}. 
 \item [AMP:] Approximate message passing algorithm from \cite{jeon2015optimality}. The algorithm was adjusted to the real-valued case 
 and was implemented with $3K$ iterations.
 \item [AMP2:] Same as the AMP algorithm but with a mis-specified SNR. The SNR in dB has an additional $\mathcal{N}(0,2)$ bias.
 \item  [SDR:] A decoder based on semidefinite relaxation implemented using a specifically tailored and efficient interior point solver \cite{luo2010semidefinite,jalden2008diversity}.
 \end{description}

 \begin{figure}[t]

  \centering
  \center{\includegraphics[width=8.5cm]{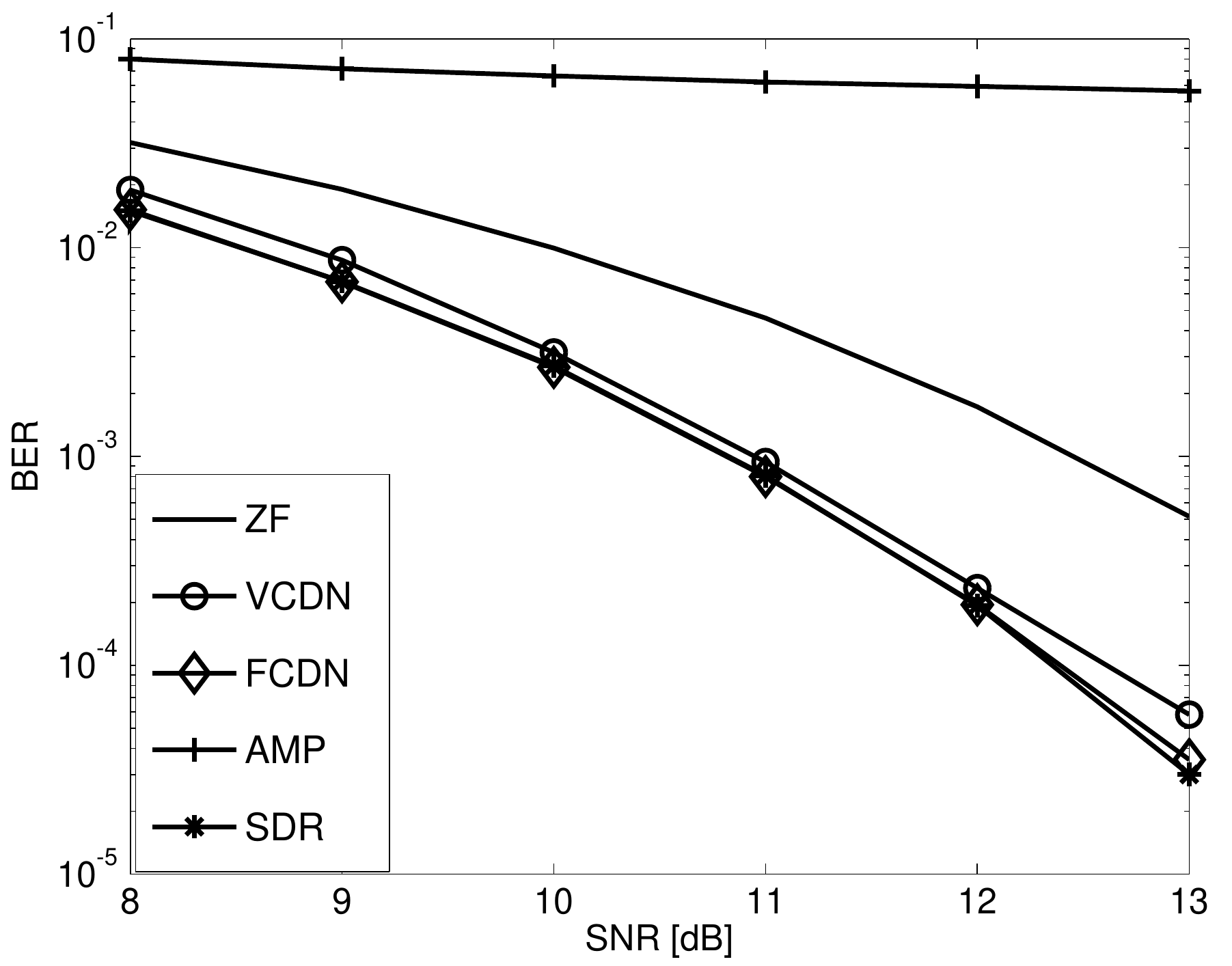}}
%
\caption{Comparison of BER performance in the fixed channel case between the detection algorithms. all algorithms were tested on the 0.55-Toeplitz channel.}
\label{fig:FC_graph}
\end{figure}

 \begin{figure}[t]

  \center{\includegraphics[width=8.5cm]{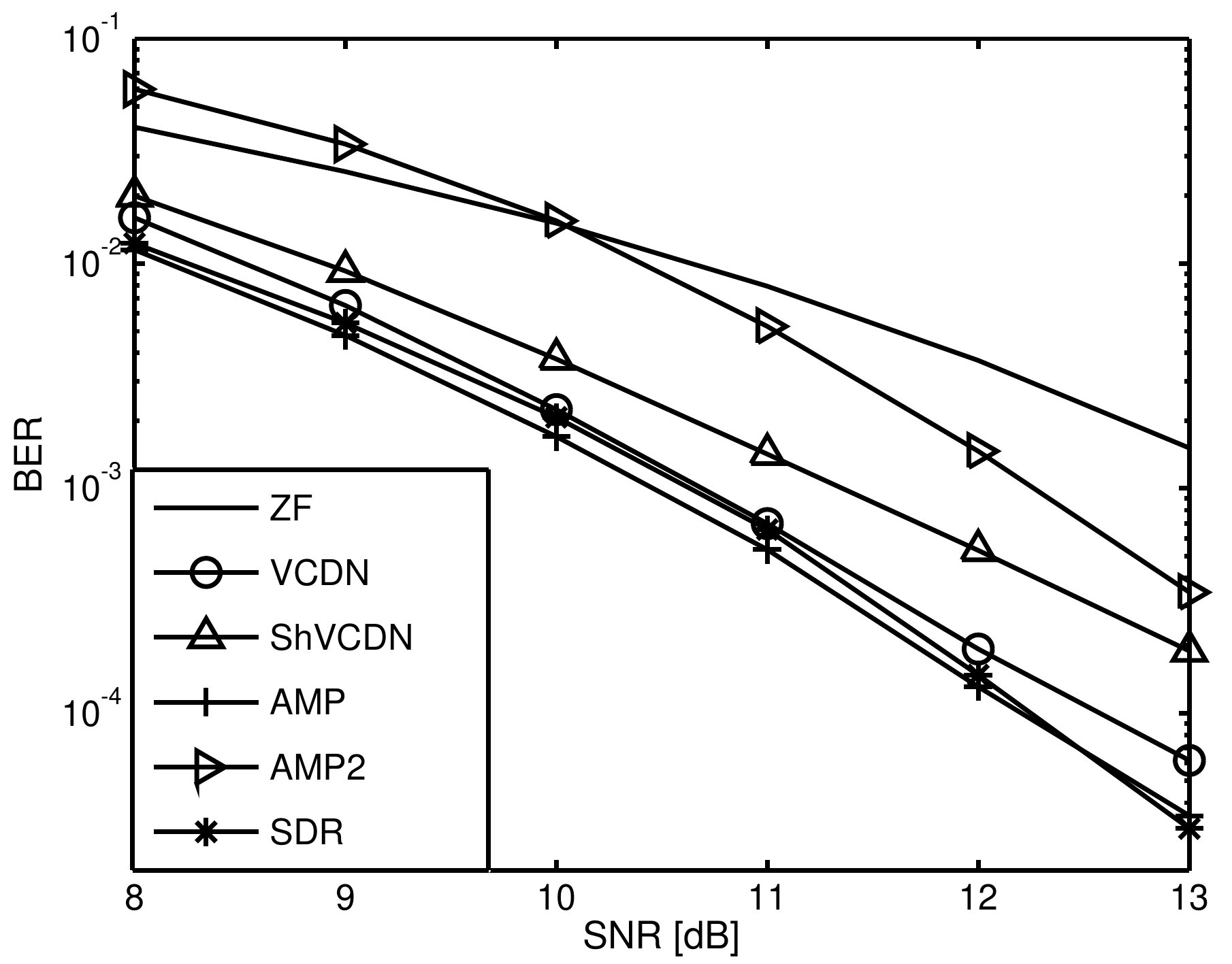}}

%
\caption{Comparison of the detection algorithms BER performance in the varying channel case . All algorithms were tested on random i.i.d. Gaussian channels.}
\label{fig:VC_graph}
\end{figure}

In our first experiment, we focused on the FC model in which the channel is known and fixed, yet challenging due to its condition number. 
Figure \ref{fig:FC_graph} shows the results of all the algorithms in this setting.  FCDN  manages to reach the accuracy rates of the computationally expensive SDR algorithm which in our simulations took 30 times longer to detect. AMP does not manage to detect with reasonable accuracy in this challenging channel. It is interesting to notice that VCDN, which was not designed for this challenging channel, also manages to achieve good accuracy. This result indicates that VCDN generalizes itself during the training phase to detect over arbitrary random channels.
 
In our second experiment which results are presented in figure \ref{fig:VC_graph}, we examine the performance in the VC model. SDR and AMP are theoretically known to be optimal in this setting, and VCDN manages to provide similar accuracy. Compared to SDR, VCDN runs 30 times faster. Compared to AMP in a scenario where the SNR values are not given accurately, we can notice a negative effect on the accuracy of the AMP, compared to VCDN that does not require any knowledge regarding the SNR.


Another important feature of DetNet is the ability to trade-off complexity and accuracy by adding or removing additional layers. In figure \ref{fig:VC_graph} we test the ShVCDN algorithm that is a shallow version on VCDN with only $K$ layers , which is much faster, but less accurate. 
Since every layer in DetNet outputs a predicted signal $\hat{\x}_k$, we can decide in real-time what layer will be the final output layer, and trade-off complexity for accuracy in real-time, without any further training.

\section{Conclusion}
In this paper we have presented deep neural networks as a general framework for MIMO detection. We have tested the performance in the fixed channel scenario over challenging channels, and in the more complicated VC scenario. The DetNet architecture we have suggested has proven to be computationally inexpensive and has near-optimal accuracy without any knowledge regarding the SNR level. The ability of DetNet to optimize over an entire distribution of channels, rather than a single or even a large-finite set of channels, makes it robust and enables implementation in systems where the channel is not fixed. Simulations show that DetNet succeeds to generalize and detect accurately over channels with different characteristics than those of the channels used in the training phase of DetNet. 
For more details, see \cite{Samuel2017learning}, where further information is presented.

\section*{Acknowledgments}
We would like to thank Prof. Shai Shalev-Shwartz for his help during the research and his insights. This research was partly supported by the Heron Consortium and by ISF grant 1339/15.

\bibliographystyle{IEEEbib}
\bibliography{main.bib}

\end{document}